\documentclass[10pt,twocolumn,letterpaper]{article}

\usepackage{cvpr}
\usepackage{times}
\usepackage{epsfig}
\usepackage{graphicx}
\usepackage{amsmath}
\usepackage{mathrsfs}
\usepackage{amssymb}
\usepackage{tabulary}
\usepackage{color}
\usepackage{url}
\usepackage{multirow}
\usepackage{stfloats}
\usepackage{makecell}
\usepackage{float}
\usepackage{overpic}
\usepackage{authblk}
\usepackage{eso-pic}

\definecolor{green}{rgb}{0, 0.5, 0}
\definecolor{turquoise}{cmyk}{0.65,0,0.1,0.1}
\definecolor{purple}{rgb}{0.65,0,0.65}
\definecolor{dark_green}{rgb}{0, 0.5, 0}
\definecolor{orange}{rgb}{0.8, 0.6, 0.2}
\definecolor{red}{rgb}{0.8, 0.2, 0.2}
\definecolor{brown}{rgb}{0.5, 0.16, 0.16}
\definecolor{red}{rgb}{1, 0.2, 0}
\newcommand{\pd}[1]{{\color{green}#1}}
\newcommand{\dc}[1]{{\color{red}#1}}

\usepackage[pagebackref=true,breaklinks=true,letterpaper=true,colorlinks,bookmarks=false]{hyperref}

\newcommand{\SUM}{\mathrm{sum}}

\def\cvprPaperID{3603} % *** Enter the CVPR Paper ID here

\pdfoutput=1 

\ifcvprfinal\fi

\cvprfinalcopy % *** Uncomment this line for the final submission

\begin{document}

% Our packages and commands
%\input{tweaks}

\title{Hausdorff Point Convolution with Geometric Priors}

%% Group authors per affiliation:
\author[1]{Pengdi Huang}
%\ead{alualu628628@gmail.com}

%% or include affiliations in footnotes:
\author[1]{Liqiang Lin}
%\ead{liniquie@gmail.com}

\author[1]{Fuyou Xue}
%\ead{fullcyxuc@gmail.com}

\author[2]{Kai Xu}
%\ead{kevin.kai.xu@gmail.com}

\author[3]{Danny Cohen-Or}
%\ead{cohenor@gmail.com}

\author[1]{Hui Huang}
%\ead{hhzhiyan@gmail.com}

\affil[1]{Shenzhen University, China}
\affil[2]{National University of Defense Technology, China}
\affil[3]{Tel Aviv University, Israel}

\maketitle

\begin{abstract}
%Learning deep features with convolution layers has immensely improved the power of deep learning. However, extending them to be applied to irregular data is non-trivial. In particular, extracting deep features on irregular and unordered point clouds remains challenging. In this paper, we introduce Hausdorff Point Convolution (HPC) to learn deep features on such point clouds. For a given point, we define a novel convolution operation, based on the Hausdorff distance between a geometric kernel shape and the neighborhood of the point. The key idea is that Hausdorff similarity measure mimics convolutional activation, and the derived HPC has notable properties: permutation-invariant, scale-invariant, and conditional rotation-invariant.

Developing point convolution for irregular point clouds to extract deep features remains challenging.
Current methods evaluate the response by computing point set distances which account only for the spatial alignment between two point sets, but not quite for their underlying shapes. 
%
%A crucial design in modern point convolution is how to compute the response between the input and a kernel point set. Existing methods evaluate the response by computing point set distance.
%The currently adopted distance measures usually account only for the spatial alignment between two point sets but not quite for their underlying shapes. 
Without a shape-aware response, it is hard to characterize the 3D geometry of a point cloud efficiently with a compact set of kernels.
In this paper, we advocate the use of Hausdorff distance as a shape-aware distance measure for calculating point convolutional responses. The technique we present, coined Hausdorff Point Convolution (HPC), is shape-aware. We show that HPC constitutes a powerful point feature learning with a rather compact set of only four types of geometric priors as kernels. We further develop a HPC-based deep neural network (HPC-DNN). Task-specific learning can be achieved by tuning the network weights for combining the shortest distances between input and kernel point sets. We also realize hierarchical feature learning by designing a multi-kernel HPC for multi-scale feature encoding. Extensive experiments demonstrate that HPC-DNN outperforms strong point convolution baselines (e.g., KPConv), achieving $2.8\%$ mIoU performance boost on S3DIS~\cite{armeni2017joint} and $1.5\%$ on SemanticKITTI~\cite{behley2019iccv} for semantic segmentation task.

\end{abstract}

\section{Introduction}

Analysis of large-scale scanned scenes is drawing increasing attention driven by advances in deep neural networks.
Point clouds acquired by common depth scanners, e.g., LiDAR, RGBD cameras, have challenging characteristics: noisy, incomplete, sparse, irregular, and unordered. These characteristics prevent applying conventional convolutions to successfully extract effective per-point features.

\begin{figure}[t]
	\centering
	\begin{overpic}[width=\columnwidth]{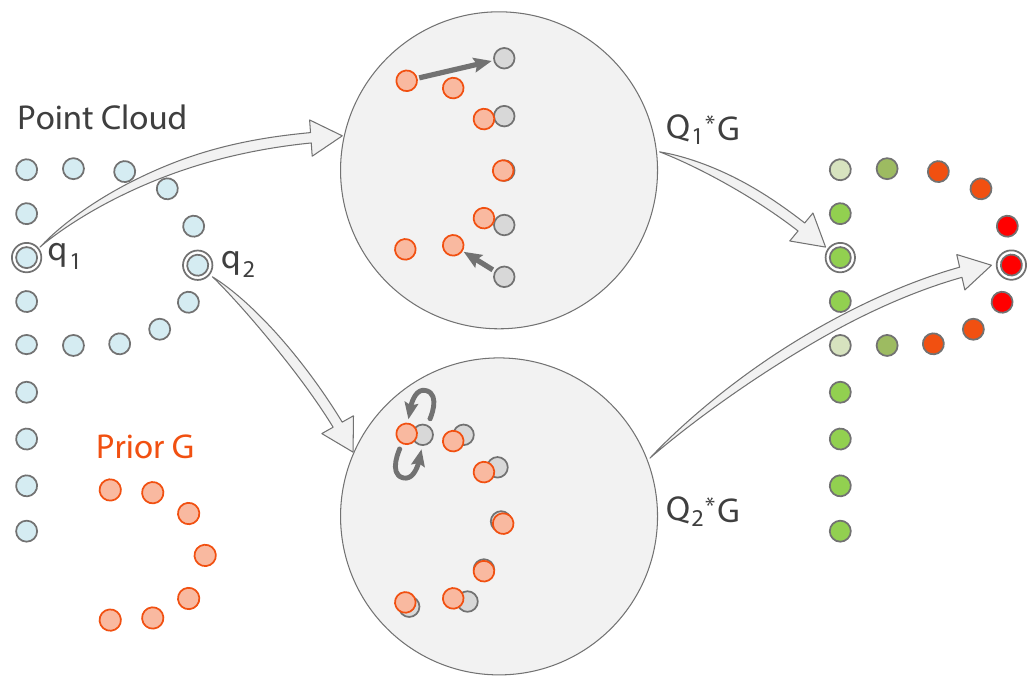}
	\end{overpic}
	\caption{In Hausdorff Point Convolution, the response of a local neighborhood ($Q_1$ or $Q_2$) to a geometric kernel ($G$) is computed based on Hausdorff distance, which is a \emph{shape-aware} distance measure. Such shape-aware convolution facilitates a powerful point feature learning with a compact set of geometric priors as convolutional kernels.
		%An illustration of Hausdorff based point convolution (HPC) on a neighboring point cloud $Q_{i}$ (gray) of query point $q_{i}$ (blue) with a prior kernel $G$ (orange). The response of HPC is a scalar mapped in color (right). \pd{HPC is implemented through the weighted Hausdorff distance approach. By weighting the shortest distance matrix between  neighboring points and kernel points with input features, the network learns to measure "shape similarity" to extract deep features.}
}
	\label{fig:teaser}
\end{figure}

Early attempts to applying convolutions on point cloud have bypassed the irregularity problem by mapping the data to predefined regular grids~\cite{Yang2018FoldingNet,rao2019spherical}.
Advanced techniques use point convolution methods based on implicit or geometric kernel functions~\cite{wu2019pointconv,lei2019octree,tatarchenko2018tangent}.
For example, KPconv~\cite{thomas2019kpconv} adopts explicit kernel point set and achieves state-of-the-art results.
%This method imitates the form of 2D grid operation and uses a pre-defined kernel function.
A crucial design choice herein is how to compute the response between the input and a kernel point sets. Existing methods usually evaluate the response by computing point set distance. The distance measures used by existing techniques only account for the spatial alignment between two point sets but not quite for their underlying shapes. Without a shape-aware response, it is hard to characterize the 3D geometry of a point cloud efficiently with a compact set of kernels.

In this paper, we advocate the use of Hausdorff distance, as a shape-aware distance measure, and show that it is particularly suited for computing convolutional response between a point cloud and a kernel point set; see Figure~\ref{fig:teaser}.
%The convolution of local neighborhood and geometric kernel is based on the Hausdorff distance measure, hence named Hausdorff Point Convolution (HPC).
We introduce a new point convolution where the response of the local neighborhood against a geometric kernel is computed based on the Hausdorff distance measure, hence named Hausdorff Point Convolution (HPC).
Since HPC is shape-aware, we are able to achieve a powerful point feature learning with a particularly compact set of only four types of geometric priors as kernels.
The key characteristic of Hausdorff distance is its robustness to noise, outliers and irregular point density, making HPC highly preferable for feature learning of typical raw scan point data.
Another property of HPC is that it is permutation-invariant and can handle data with arbitrary scales.
Furthermore, rotation invariance can be achieved by adopting a rotationally-symmetric kernel shape.

We developed a HPC-based deep neural network, referred to as HPC-DNN, by adopting KPConv~\cite{thomas2019kpconv} as the basic architecture.
To make HPC operations learnable within the network, we decompose HPC into a \emph{distributive function} and a \emph{shortest distance matrix} between the point sets of the local neighborhood and a geometric kernel.
The distributive function is then used to aggregate the shortest distances weighted by the learnable parameters and the input features, which is essentially a weighted Hausdorff distance.
Task-specific learning is achieved by tuning the network weights for shortest distance aggregation.
To realize hierarchical feature learning, we design multi-kernel HPCs for multi-scale feature learning.

We have implemented HPC-DNN with PyTorch and evaluated it on several classic tasks of point cloud processing and understanding, i.e., semantic object detection and segmentation on point scanned scenes.
Experiments show that HPC-DNN outperforms strong baseline point convolutions (e.g., KPConv) significantly. In particular it attains $2.8\%$ mIoU improvement on S3DIS~\cite{armeni2017joint} and $1.5\%$ on SemanticKITTI~\cite{behley2019iccv} for semantic segmentation task.

\section{Related Work}

The analysis of unstructured point cloud is widely regarded as a difficult and ill-posed problem. In particular, feature extraction is a fundamental and important one~\cite{steder2010narf,rusu2010fast,salti2014shot}. 
Recently, with the emergence of neural networks, new methods for point cloud analysis have been introduced ~\cite{qi2017pointnet,qi2017pointnet++,li2018pointcnn}.
These methods build upon the ability of neural networks to learn from data.
The key challenge in deploying neural networks is that point clouds are irregular and unordered, so conventional convolutions cannot be adopted. Previous works can be roughly categorized into three: i) Grid-based, ii) Implicit, and iii) Explicit point convolution. We briefly discuss them as follow.

\paragraph{Grid-based convolution on point clouds.}
To bypass the data irregularity, these grid-based methods project or transform the data into regular grids, on which traditional convolution methods can be applied~\cite{Su15MV,maturana2015voxnet}. These methods are mainly designed for individual objects. For example, in SFCNN~\cite{rao2019spherical}, point clouds are projected onto a grid sphere. The local and global features are then learned by a multi-layer perception (MLP) architecture.
FoldingNet~\cite{Yang2018FoldingNet} proposes a two-step-folding operation to construct a mapping between point set and 2D grid.
To obtain higher-level features, some researches adopt the VoxelNet framework.
In~\cite{ye2019sarpnet}, point clouds are voxelized and a shape attention regional proposal network is trained to learn the spatial occupancy of objects in horizontal and vertical directions. Grid-based convolution is limited by resolution due to heavy computation cost, which might be relieved with efficient data structure~\cite{riegler2017octnet,klokov2017escape}. Nonetheless, this thread of methods are, in general, sensitive to noise and tend to overlook small-scale, yet visually meaningful shape details. For larger and more complex scenes, specific designs need to deal with incomplete data and outliers.

\paragraph{Implicit point convolution.}
The pioneering work of PointNet~\cite{qi2017pointnet}, opened an avenue of works focusing on direct convolution on 3D point clouds, which either aim to improve the neighborhood structure~\cite{qi2017pointnet++,graham20183d,su2018splatnet}, or to enhance the convolutional filters~\cite{simonovsky2017dynamic,hermosilla2018monte,xu2018spidercnn}.
Atzmon et al.~\cite{atzmon2018point} propose a unique volume-based point convolution, which consists of two operators, i.e., extension and restriction, mapping point cloud functions to volumetric ones and vise-versa.
Hu et al.~\cite{hu2020randla} propose a location spatial encoding block to group relative coordinates and input features, and then extract the neighboring features. Li et al.~\cite{li2018pointcnn} present PointCNN, which uses MLP to learn an $X$ matrix to canonize the point cloud features, thus permutation- invariant and offering hierarchical convolutions.
PointConv~\cite{wu2019pointconv} constructs a location related weight function for continuous convolution, and re-weight the weight function through a learned point density factor.
Tatarchenko et al.~\cite{tatarchenko2018tangent} propose a TangentConv that uses Gaussian kernels as implicit kernel metrics functions.
DGCNN~\cite{wang2019dynamic} adopts an EdgeConv, which constructs a local graph on neighboring points. Graph information function is instantiated by a fully connected layer.
All these methods locally organize the geometric features~\cite{liu2019relation,2019Modeling}, and then use MLP to obtain the final high-level features,
referred to as implicit point convolution. These implicit convolutions learn permutation or rotation invariance by MLP layers, thus sensitive to training data quality and the network training convergence.

\paragraph{Explicit point convolution.}
Deep feature extraction by analyzing local neighborhoods has received intensive attentions, but no much focus has been given to the development of explicit reference shapes to promote the deep feature expressiveness.
Recently, KPConv~\cite{thomas2019kpconv} introduces a point kernel based method for point cloud convolution operations, which achieved state-of-the-art performances on classic point cloud datasets. However, this method simply applies the form of a two-dimensional grid operation and uses a prescribed kernel function.
JSENet~\cite{thomas2019kpconv} adopts KPConv as its backbone network for point convolution, and proposes a deep network that fuses the region and edge information for joint learning of semantic segmentation and edge detection.
Taking a perspective of blending both geometry and topology, we define a permutation-invariant and scale-invariant geometric convolution, offering the analysis of unordered point clouds. By combining feature mapping and geometric convolution, we construct a convolutional neural network that learns to weight the input shape feature. Moreover, based on the diversity of shapes, the multi-kernels are designed to extract hierarchical features jointly.

\section{Method}

\begin{figure*}[t]
	\centering
	\begin{overpic}[width=\textwidth,tics=5]{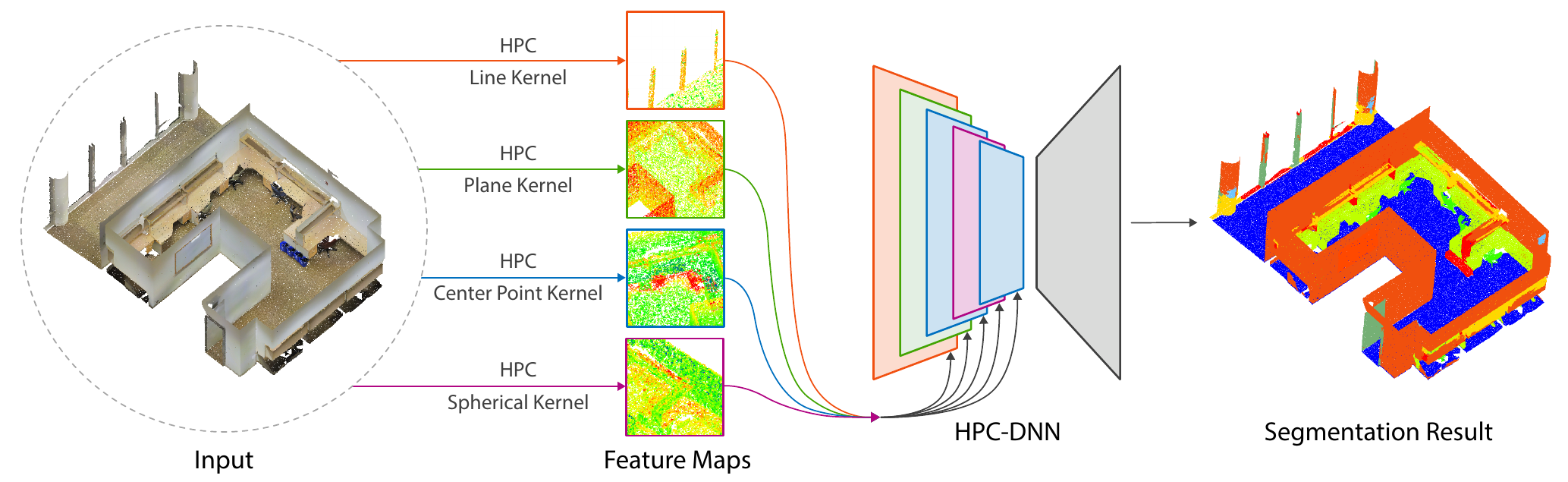}
	\end{overpic}
	\caption{
		An overview of the proposed HPC-DNN. The HPC-DNN adopts a multi-kernels Hausdorff based point convolution (HPC), and extracts hierarchical features from the large-scale input point cloud scene. Then, multi-scale convolution responses of different kernels are merged in each layer to implement scene semantic segmentation.}
	\label{fig:overview}
\end{figure*}

\paragraph{Overview}
We propose Hausdorff point convolution (HPC), a new point convolution based Hausdorff distance.
In HPC, there are four types of convolutional kernels whose parameters are learnable with downstream tasks such as point cloud segmentation; see Figure~\ref{fig:overview}. The feature response between input points and kernel points is computed based on Hausdorff distance which essentially measures the similarity between the two point sets.
The feature responses of multiple kernels at the same scale (layer) are combined to form a powerful representation.
Deep networks are formed by stacking multi-scale HPCs which learn hierarchical feature representations and finally produce an output.
%Finally, the network integrates the extracted features to predict the label of each point.
%In the following, Sec.~\ref{HPC_def} introduces the definition of Hausdorff convolution, Sec.~\ref{HPC_layer} introduces the Hausdorff convolution layer, and Sec.~\ref{HPC_multi} introduces the Hausdorff convolution with multiple kernels.

\subsection{Definition of HPC}
\label{HPC_def}
Given an input feature vector $F=f(v)$ and a kernel vector $G=g(v)$, the discrete convolution writes as:
\begin{equation}
F*G = \sum_{u\in{U}} f(u)g(v-u) = \langle f(u), g(v-u)\rangle,
\end{equation}
where $*$ denotes convolution operation, and $\langle\cdot,\cdot\rangle$ is vector inner product. Convolution is essentially a ``sliding inner product'' between the input feature $f(u)$ and the flipped kernel $g(v-u)$.
%The convolution of and $g(u)$ is equivalent to inner product of $f(u)$ and, where $g(v-u)$ is the flip of $g(u)$.
Inner product measures the similarity between two vectors. The convolution response thus reflects a ``sliding similarity'' between the ``shape'' of $F$ and $G$.

In 2D convolutional layers of a CNN, the similarity is measured between a 2D feature map and a 2D filter for feature extraction. We hope to extend this concept of convolution to deep feature learning of 3D point clouds. We give a generalized definition of \emph{point convolution}. Given a point $p_{o} \in P$ and its neighboring point set $Q$, and a given kernel point set $G$, we use a function $T$ to calculate the similarity response $\hbar$ between each other:
\begin{equation}
\hbar = T(Q,G), Q = \{p_{i} \mid \| p_{o} - p_{i} \| \le r\},
\end{equation}
where $r$ is the query radius.
%Assuming that the point cloud is a sampling of implicit shapes, the similarity calculation is equivalent to the shape matching.
Hausdorff distance can be used for computing shape similarity.
Given the neighboring point set $Q$ and a kernel point set $G$, the Hausdorff distance $H(Q,G)$ is adopted as the convolution function $T$:
\begin{equation}
H(Q,G) = \max (h(Q,G),h(G,Q)),
\label{con:rawhausdorff}
\end{equation}
where $h(G,Q)$ and $h(Q,G)$ are called the narrow Hausdorff distance. Their formula is defined as follows:
\begin{equation}
\begin{split}
h(G,Q) &= \max_{g\in{G}}\min_{q\in{Q}}\| g - q \|, \\
h(Q,G) &= \max_{q\in{Q}}\min_{g\in{G}}\| g - q \|,  \label{con:disptoq}
\end{split}
\end{equation}
where $\| g - q \|$ is the Euclidean distance between point $g$ and $q$. Obviously,  $H(G,Q) = H(Q,G) \leq r$, but $h(G,Q) \neq h(Q,G)$ in general.

If the kernel point set $G$ is defined in a spherical space, $G = \{g_{i} \mid \| g_{i} \| \le r\}$, the neighborhood point set $Q$ and the kernel point set $G$ do not necessarily need a rigid registration, which means that Hausdorff distance measurement can be performed directly on the two point sets.

\begin{figure}
	\begin{overpic}[width=\columnwidth,tics=5]{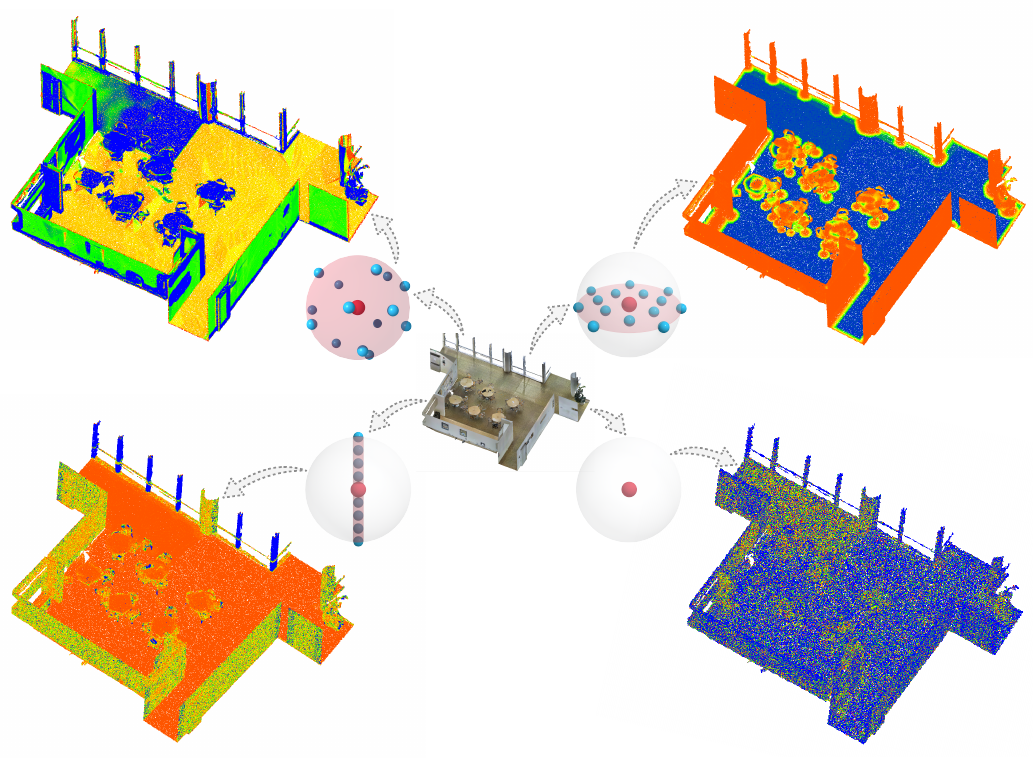}
	\end{overpic}
	\caption{
		Hausdorff distance metric on an indoor scene. The point color (dark to bright) indicates the Hausdorff distance value (low to high) between neighborhood points and the given kernel.}
	\label{fig:hausresponse}
\end{figure}

In fact, the Eq.~\eqref{con:rawhausdorff} satisfies the important properties of point cloud convolutions described in~\cite{li2018pointcnn}. For $Q = \left[q_{1},\ldots, q_{n}\right]$ and its permuted counterpart $Q' = \left[q'_{\pi_1},\ldots, q'_{\pi_n}\right]$, it has \emph{point permutation invariance}: $H(Q,G) = H(Q',G)$. Besides, it is \emph{scale invariance} after normalization as $H(Q,G) = H(rQ,rG)/r$ with the query radius $r$ being a given constant. Since $H(G,Q) \leq r$, $1/r$ is then the normalization factor. This means that a kernel shape can calculate neighboring shape responses at any scale or query radius. Figure~\ref{fig:hausresponse} shows the result of calculation using the Eq.~\eqref{con:rawhausdorff} on a 3D indoor scene. It can be seen that the vertical line kernel causes a notable response on the pole-like structures, while the plane kernel has a notable response on the ground. This result is very similar to the neuron activation of feature maps in 2D convolutional neural networks.

\begin{figure*}
	\begin{overpic}[width=1.0\textwidth]{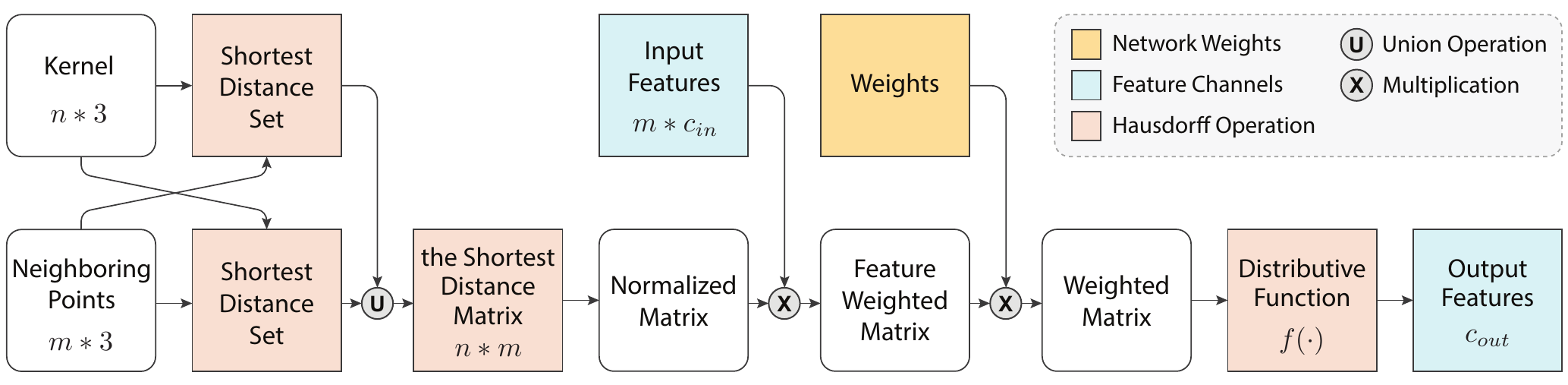}
	\end{overpic}
	\caption{
	      A point convolutional layer of the proposed HPC-DNN. The inputs are a query point and its neighboring points with corresponding features and a given kernel points, the output is a feature vector of query point.}
	\label{fig:convlayer}
\end{figure*}

\subsection{HPC Layer}
\label{sec:HPC_layer}
As a network layer, Hausdorff convolution layer should contain learnable weights which can be optimized via back propagation. Instead of computing the Hausdorff response directly, we opt to assign a weight corresponding to each shortest distance, and automatically adjust the length according to the attitude and distribution of neighboring points with input features. We split the Hausdorff distance operator into two parts: the shortest distance set and the distributive function (see definition below). The distance between a neighboring point $q_{i}$ and a kernel point set $G$ is defined as $d(q_{i},G) = \min_{g\in{G}}\|g - q_{i}\|$, and similarly the distance between a kernel point $g_{i}$ and a neighboring point set $Q$ is $d(g_{i},Q) = \min_{q\in{Q}}\|g_{i} - q\|$. Let set $D_{g}$ represent the set of distances from each point in $G$ to the set $Q$ as $D_{g} = \left[d(g_{1},Q),\ldots, d(g_{n},Q)\right]$, and $D_{q}$ be $D_{q} = \left[d(q_{1},G),\ldots, d(q_{m},G)\right]$, where $n=|G|$ and $m=|Q|$. Therefore, the Hausdorff convolution between $Q$ and $G$ can be written as:
\begin{equation}
\begin{split}
	H(Q,G) &= \max (\max (D_{g}),\max (D_{q})) \\
	&=  \max ([D_{g},D_{q}]).
\end{split}
\end{equation}

Let us name a function $f$ satisfying $f(\{f(D_{g})\} \cup \{f(D_{q})\}) = f(D_{g} \cup D_{q})$ a \emph{distributive function}. $\left[D_{g}, D_{q}\right] = D_{g} \cup D_{q}$ is the shortest distance set. It can be seen that the  $\max$ function is a distributive function. However, the $\max$ operator is sensitive to outliers. To overcome this issue, many improved variants of Hausdorff distance has been proposed~\cite{dubuisson1994modified}. Among the variants, we found the Hausdorff distance in the cumulative form~\cite{jesorsky2001robust} is more appropriate for pattern matching. This modified Hausdorff distance is:
\begin{equation}
h(Q,G) = \sum_{q\in{Q}}\min_{g\in{G}}d(g,q). \\
\end{equation}
In this case, Eq.~\eqref{con:rawhausdorff} and Eq.~\eqref{con:disptoq} would also adopt accumulation operation instead of max operation. Let $\SUM$ indicate the accumulation of elements in a set, it can be:
\begin{equation}
\begin{split}
H(Q,G) &= \SUM (\SUM (D_{g}), \SUM (D_{q})) \\
&=  \SUM ([D_{g},D_{q}]).
\end{split}
\end{equation}
Obviously, the accumulation function is also a distributive function. We construct a shortest distance matrix $D_\text{min}$ to store the shortest distance set $D_{g} \cup D_{q}$:
\begin{equation}
D_\text{min}(i,j) =\left\{
\begin{aligned}
\|q_{i} - g_{j} \|,& &\|q_{i}-g_{j}\|\in D_{g} \cup D_{q} \\
0,& &\text{otherwise}
\end{aligned}
\right.
\end{equation}
where $1\leq i \leq m$ and $1 \leq j\leq n$. The size of $D_\text{min}$ is $n \times m$.

The point convolutional network layer is to weight the elements in $D_\text{min}$, and the input feature of each point can also be considered as a distance weighting constant. Inspired by KPConv~\cite{thomas2019kpconv}, we formulate the computation of Hausdorff convolution based on the shortest distance matrix $D_\text{min}$. For a input features $F_\text{in} \in {\mathbb{R}^{m \times c_\text{in}}}$, and an output features $F_\text{out} \in{\mathbb{R}^{c_\text{out}}}$, the convolution formula of the point convolutional layer is:
\begin{equation}
F_\text{out} = f(D_\text{min}F_\text{in}W), \\
\end{equation}
where $W$ is the weight matrix with a size of $n \times  c_\text{in} \times  c_\text{out}$. It maps the features from input channel number $c_{in}$ to the output channel number $c_{out}$. The shortest distance matrix is a sparse matrix: $\| D_{min} \|_{0} \leq n + m$. Therefore, multiplying $D_\text{min}$ by the weight matrix $W$ approximates weighting each shortest distance as $w_{ji}D_\text{min}(i,j)$. Since distributive functions including $\max$, $\SUM$, $\min$ are all differentiable, the Hausdorff convolutional layer is also differentiable. Figure~\ref{fig:convlayer} shows the architecture of a HPC layer. As can be seen, the shortest distance matrix is normalized before weighting. Normalization is to remove the scale factor: $\bar{D}_\text{min}(i,j) = D_\text{min}(i,j)/r$. In addition, since the Hausdorff distance measures the maximum dissimilarity between two shapes, for similarity response, the nonzero value should take $1 - \bar{D}_\text{min}(i,j)$.

Hausdorff convolution is a general form of point cloud convolution operation. From the view of Hausdorff convolution, KPConv~\cite{thomas2019kpconv} is equivalent to computing a one-way similarity by using a self-defined measurement function (kernel function). PointConv~\cite{wu2019pointconv} and RSCNN~\cite{liu2019relation} are equivalent to matching with a kernel containing only one point at the center. PointConv uses density information to weight distance, while RSCNN adopts maximum distance.

\subsection{HPC with Multiple Kernels}
\label{HPC_multi}
Similar to common CNNs, we require multiple kernels (the point cloud $G$) to facilitate a powerful feature learning. This involves the design of kernel point generation and multi-kernel network structure.

\paragraph{Kernel point generation}
The choice of the shape of kernel point clouds is important. They should be geometric primitives (e.g., planar patches, cylindrical patches, quadric patches, etc.), and meanwhile they should be ideally omnipresent on the 3D surface of common objects and scenes. In this work, we find that the following four types of kernel shapes suffice: points, lines, planes, and spherical patches, encompassing primitives from 0D to 3D. Further, we find that Hausdorff response is more robust if $G$ has a rotationally symmetric structure:
\begin{equation}
H(Q,G) = H(RQ,G),
\end{equation}
where $R\in SO(3)$ is a rotation transformation being applied to the neighborhood point set $Q$. Apparently, point and spherical kernels bring rotation invariance to the calculation, which qualifies Hausdorff distance for retrieving the same kernel in varying poses. To form the kernel point sets, we sample the primitive shapes using the farthest point sampling (FPS) algorithm. Figure~\ref{fig:hausresponse} shows the four kernel shapes adopted in our method.

\begin{figure}[t]
	\begin{overpic}[width=\columnwidth,tics=5]{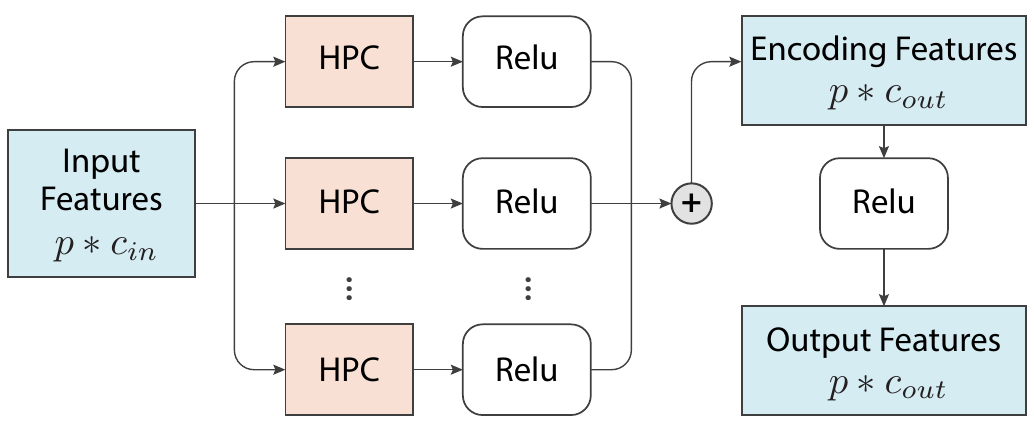}
	\end{overpic}
	\caption{The architecture of multi-kernel Hausdorff point convolution layer. For a query point and its input feature $F_\text{in}$, output feature $F_\text{out}$, the convolution results $c_\text{out}$ of $K$ kernels are linearly accumulated.}
	\label{fig:multiHPC}
\end{figure}

\paragraph{Multi-kernel network structure}
With multiple kernel point sets, we hope that each convolution feature component contributes to the final encoding feature. Therefore, a multi-kernel feature encoding $\tilde{F}_{out} \in{\mathbb{R}^{c_{out}}}$ groups $K$ Hausdorff point convolution result $F_{out} \in{\mathbb{R}^{c_{out}}}$:
\begin{equation}
\tilde{F}_{out} = ReLU(\sum_{k=1}^K ReLU(F_{out}(G_{k}))),
\end{equation}
where $K$ is the number of kernels and $ReLU$ is the activation function of rectified linear units. Figure~\ref{fig:multiHPC} shows the architecture of multi-kernel Hausdorff point convolution. Different kernel features are grouped by summing operation. Compared with concatenation operation, the parameter amount of summing is smaller and the common part between features can be mutually enhanced.

\section{Results and Evaluation}

Our network, HPC-DNN, is implemented based on KP-FCNN~\cite{thomas2019kpconv}.
The network architecture consists of two parts: an encoder and a decoder.
The encoder contains five convolutional layers.
Each convolutional layer contains two HPC layers or two multi-kernel HPC layers. For a feature of dimension $c_{in}$, the input and output dimension of the first convolution operation is $c_{in}$ and $2*c_{in}$, respectively. The input and output dimensions of the second convolution layer are both $2*c_{in}$.
The query radius $r$ of point neighborhood doubles for every layer.
The decoder, following the encoder, acts as deconvolution.
Deconvolution implements point feature propagation based on nearest neighbor upsampling.
We retain the shortcut structure of KP-FCNN in convolutional layers.
For multi-kernel HPCs, the shortcut structure is removed for building a concise convolutional layer.

We implement our network with PyTorch and perform training/testing on a server with 22 Intel 2.20GHz Intel(R) Xeon(R) E5-2699 CPU and 9 Quadro GV100.
%The motivation of our method is to perform automatic feature extraction on the collected real point cloud data to solve the actual problem of scene semantic segmentation.
%The point cloud semantic segmentation is to predict the semantic label of each point data.
We conduct scene semantic segmentation tasks on large-scale point clouds of indoor and outdoor scenes for evaluation.

\subsection{Segmentation on S3DIS}
For indoor scene segmentation, we use the public indoor point cloud dataset S3DIS~\cite{armeni2017joint} to compare our proposed HPC-DNN with KPConv~\cite{thomas2019kpconv}, JSENet~\cite{hu2020jsenet} and other state-of-the-art point convolution methods.
S3DIS contains $271$ indoor rooms encompassing offices, corridors, etc.
The 3D data were acquired with RGBD scanner, and the dense point cloud data is associated with RGB color information.
Following the convention, we use the set of Area-1 to Area-4 and Area-6 for training, and the set of Area-5 for testing.
Segmentation accuracy is measured by mean of intersection over union (mIoU).

Table~\ref{tab:s3dis} shows that our HPC-DNN exhibits a significant performance improvement for scene segmentation compared to other baselines. We achieve a $66.7\%$ for single kernel HPC and $68.2\%$ for multi-kernel HPC, which are both higher than KPConv~\cite{thomas2019kpconv} under the same configuration. In particular, the results of multi-kernel HPC-DNN achieves the state-of-the-art performance among all methods. Due to the high scanning quality of S3DIS scenes, the geometry of the objects is relatively complete. The multi-kernel HPC-DNN can effectively capture and enhance the geometric features of semantic targets.

\begin{table}
	\centering
	
	\caption{Semantic scene segmentation results on S3DIS.}
	\label{tab:s3dis}
	\setlength{\tabcolsep}{7mm}{
	\begin{tabular}{ccc}
		\hline
		Method                                      &            mIoU \\
		\hline
		\hline
		TangentConv~\cite{tatarchenko2018tangent}   &            52.6 \\		
		PointNet++~\cite{qi2017pointnet++}          &            53.4 \\
		DGCNN~\cite{wang2019dynamic}                &            56.1 \\
		PointCNN~\cite{li2018pointcnn}              &            57.3 \\
		PointNet~\cite{qi2017pointnet}              &            57.8 \\
		ParamConv~\cite{wang2018deep}               &            58.3 \\
		PointWeb~\cite{zhao2019pointweb}            &            60.3 \\
		HPEIN~\cite{jiang2019hierarchical}          &            61.9 \\
		SPGraph~\cite{landrieu2018large}            &            62.1 \\
		MVPNet~\cite{jaritz2019multi}               &            62.4 \\
		Point2Node~\cite{han2020point2node}         &            63.0 \\
		MinkowskiNet~\cite{choy20194d}              &            65.4 \\		
		KPConv~\cite{thomas2019kpconv}              &            65.4 \\
		JSENet~\cite{hu2020jsenet}                  &            67.7 \\
		\hline
		HPC-DNN                                     &            66.7 \\
		Multi-kernel HPC-DNN                        &   \textbf{68.2} \\
		\hline
	\end{tabular}
	}
\end{table}

\begin{figure*}[t!]
\centering
\begin{overpic}
[width=1.0\textwidth]{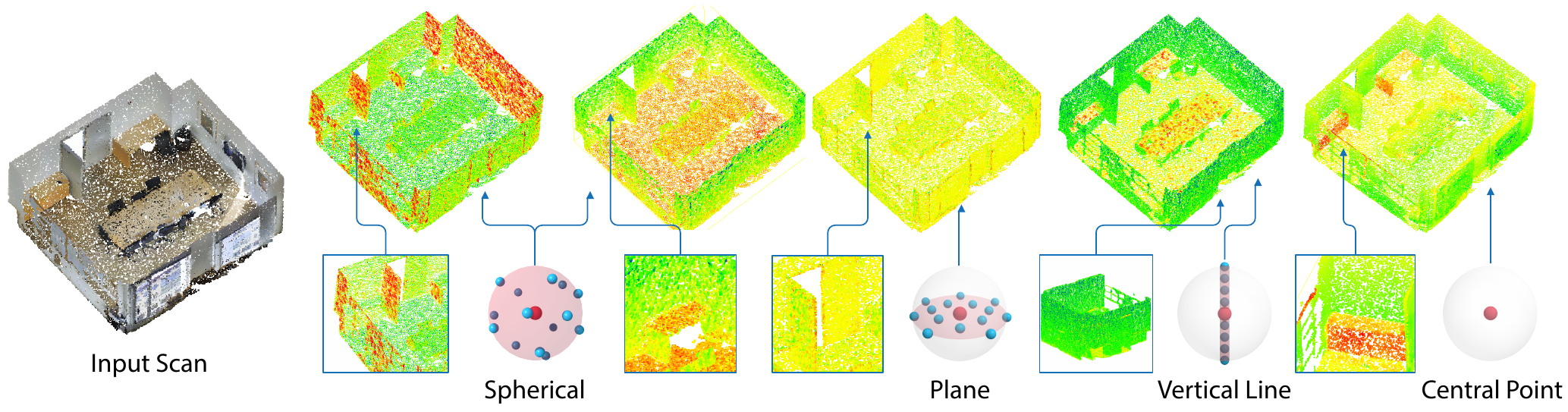}
%\put(2,18){\small (a)}
%\put(21,23){\small (b)}
%\put(38,23){\small (c)}
%\put(55,23){\small (d)}
%\put(70,23){\small (e)}
%\put(86,23){\small (f)}
\end{overpic}
\caption{Visualization of feature maps on point clouds after point convolution with different kernels.}
\label{fig:FeatureMaps}
\end{figure*}

\subsection{Segmentation on SemanticKITTI}
SemanticKITTI~\cite{behley2019iccv} is a dataset of large-scale outdoor point clouds built based on the KITTI vision benchmark~\cite{geiger2012cvpr}.
The data sequences in SemanticKITTI are composed of continuous frames of point cloud captured by LiDAR scanners without color information.
Although a single frame of point cloud contains about $100$K points, they are usually scattered in a space of $160 \times 160$ meters. Therefore, the point clouds are very sparse.
Besides, due to the round scanning nature of LiDAR, the density of the point clouds is heavily anisotropic, which further increase the difficulty of processing and understanding.
We use sequences 0--7 and 9--10 for training, and sequence 8 for testing.
We again use mIoU to evaluate the prediction accuracy over the 19 categories in the dataset.
Our method is compared to KPConv~\cite{thomas2019kpconv}, PolarNet~\cite{zhang2020polarnet} and other baselines. The hyper-parameters, e.g., the query radius of each layer and the number of sampled points, are kept consistent across different deep networks.

Table~\ref{tab:semantickitti} shows the scene segmentation result of our networks and other baselines.
HPC-DNN and Multi-kernel HPC-DNN get an mIoU of $59.6\%$ and $60.3\%$, respectively, outperforming the state of the art KPConv. Although the point clouds of this dataset contain much missing data, HPC is still able to capture as much geometric information as possible, leading to improved understanding results.

\begin{table}
	\centering
	\caption{Semantic scene segmentation results on SemanticKITTI.}
	\label{tab:semantickitti}
	\renewcommand\arraystretch{1.2}
	\setlength{\tabcolsep}{7mm}{
	\begin{tabular}{cc}
		\hline
		Method                                      &            mIoU \\
		\hline
		\hline
		PointNet++~\cite{qi2017pointnet++}          &            20.3 \\
		RSCNN~\cite{landrieu2018large}              &            47.2 \\
		PolarNet~\cite{zhang2020polarnet}           &            58.2 \\
		KPConv~\cite{thomas2019kpconv}              &            58.8 \\
		\hline
		HPC-DNN                                     &            59.6 \\
		Multi-kernel HPC-DNN                        &   \textbf{60.3} \\
		\hline
	\end{tabular}
	}
\end{table}

\subsection{Ablation Study}
\paragraph{Distributive functions.}
We analyze the performance of our HPC-DNN with different kinds of distributive functions (see definition in Section~\ref{sec:HPC_layer}).
The experiment was carried out on the S3DIS dataset using a spherical kernel, with $\max$, $\min$, and $\SUM$ functions used, respectively.
As shown in Table~\ref{tab:ffunction},
the accuracy of the $\SUM$ function is $65.41\%$ which is better than $65.04\%$ of $\min$ and $65.01\%$ of $\max$.
Although these three functions are all metric functions recommended by Dubuisson and Jain~\cite{dubuisson1994modified}, the cumulative form of $\SUM$ is more preferable in overcoming the outlier issue.

\begin{table}
	\centering
	\caption{The effect of different distributive function $f$.}
	\label{tab:ffunction}
	\renewcommand\arraystretch{1.2}
	\setlength{\tabcolsep}{7mm}{
	\begin{tabular}{lc}
		\hline
		$f$ function                 &       mIoU \\
		\hline
		\hline
		$\max$ function           &       65.41 \\
		$\min$ function           &       65.05 \\
		$\SUM$ function           &      \textbf{66.65} \\
		\hline
		
	\end{tabular}
	}
\end{table}

\paragraph{Shortest distance matrix.}
In Section~\ref{sec:HPC_layer}, we use shortest distance matrix to capture the correlation between the kernel and target shapes. Shortest distance matrix is a sparsification of neighborhood distance matrix with only the shortest distances kept and the rest set to $0$.
Here, we make a comparison between the shortest distance matrix and neighborhood distance matrix using S3DIS.
%We take the pairwise distances between kernel points and neighboring points into the matrix form to replace the shortest distance matrix for comparison.
Table~\ref{tab:shortestmat} shows that the performance of shortest distance matrix is much better than that of neighborhood distance matrix.
Although the shortest distance matrix contains less information than the neighborhood distance matrix,
the results demonstrate that the former effectively retains the key information of point neighborhood and avoids the interference of redundant data.

\begin{table}
	\centering
	\caption{Analysis of the shortest distance factor: with and without constructing the shortest distance matrix.}
	\label{tab:shortestmat}
	\renewcommand\arraystretch{1.2}
	\setlength{\tabcolsep}{7mm}{
	\begin{tabular}{lc}
		\hline
		Distance type                            &    mIoU \\
		\hline
		\hline
		All distances between $G$ and $Q$    &    61.84 \\
		The shortest distance                &     \textbf{66.65} \\
		\hline
		
	\end{tabular}
	}
\end{table}

\subsection{Impact of Kernels}

\paragraph{Kernel shape.}
We tested the effect of different kernel shapes over the S3DIS dataset.
The kernel shapes being tested include 3D sphere, 2D plane, 1D straight line, and a center point.
The number of points in each kernel is $15$, and all the points are generated by sampling the parametric shape with farthest point sampling.
The results are shown in the Table~\ref{tab:kernelshapes}.
It can be seen that spherical kernel achieves the best performance of $66.65\%$. Planar kernel, straight line kernel, and center point kernel obtain $64.02\%$, $62.16\%$ and $44.70\%$, respectively.
The good performance of spherical kernel is mainly attributed to invariant to rotation in calculating geometric distances.
Since the point kernel concentrates at the center, it is less informative.

\begin{table}
	\centering
	\caption{The impact of kernels in different shapes.}
	\label{tab:kernelshapes}
	\renewcommand\arraystretch{1.2}
	\setlength{\tabcolsep}{7mm}{
	\begin{tabular}{lc}
		\hline
		Kernel prior shape      &       mIoU \\
		\hline
		\hline
		Sphere                  &       \textbf{66.65} \\
		Plane                   &      64.02 \\
		Straight Line           &      62.16 \\
		Center point            &      44.70 \\
		\hline
	\end{tabular}
	}
\end{table}

\paragraph{Number of kernels.}
We also analyze the impact of the number of kernels using, again, S3DIS. The numbers of kernels tested are $K=1$, $K=2$, and $K=4$. The results are shown in the Table~\ref{tab:mutikerneks}.
As can be seen, the performance of network with multi-kernels significantly improves over the single kernel version.
Among multi-kernel versions, two-kernel leads to a performance improve of $1.2\%$, and four-kernel $1.6\%$.
This shows that the features extracted by the multi-kernel HPC can effectively complement to each other, and provide more powerful feature description for the point clouds.
Of course, multi-kernel features contain redundancy, which explains the slower performance improvement when four kernels are used.

\begin{table}
	\centering
	\caption{The results of different numbers of kernels.}	
	\label{tab:mutikerneks}
	\renewcommand\arraystretch{1.2}
	\setlength{\tabcolsep}{7mm}{
	\begin{tabular}{lc}
		\hline
		The number of Kernel        &       mIoU \\
		\hline
		\hline
		Kernel (Sphere)                      &       66.65 \\
		Kernels (Sphere  + Plane)             &       67.85 \\
		Kernels (All Four)                         &        \textbf{68.23} \\
		\hline
		
	\end{tabular}
    }
\end{table}

\begin{figure*}[bht]
\centering
\begin{overpic}
[width=0.97\textwidth]{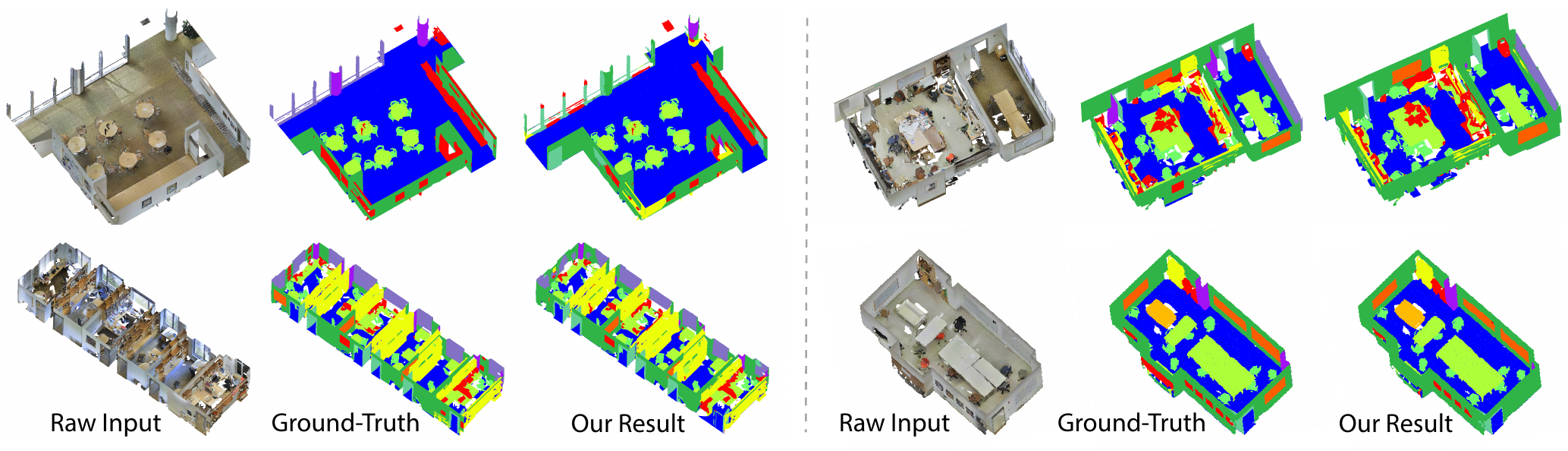}
%\put(3,1){\small (a)}
%\put(20,1){\small (b)}
%\put(37,1){\small (c)}
%\put(55,1){\small (d)}
%\put(72,1){\small (e)}
%\put(88,1){\small (f)}
\end{overpic}
\caption{A gallery of HPC-DNN segmentation results on the S3DIS dataset.
}
\label{fig:s3disgallery}
\end{figure*}

\begin{figure*}[bht]
\centering
\begin{overpic}
[width=0.97\textwidth]{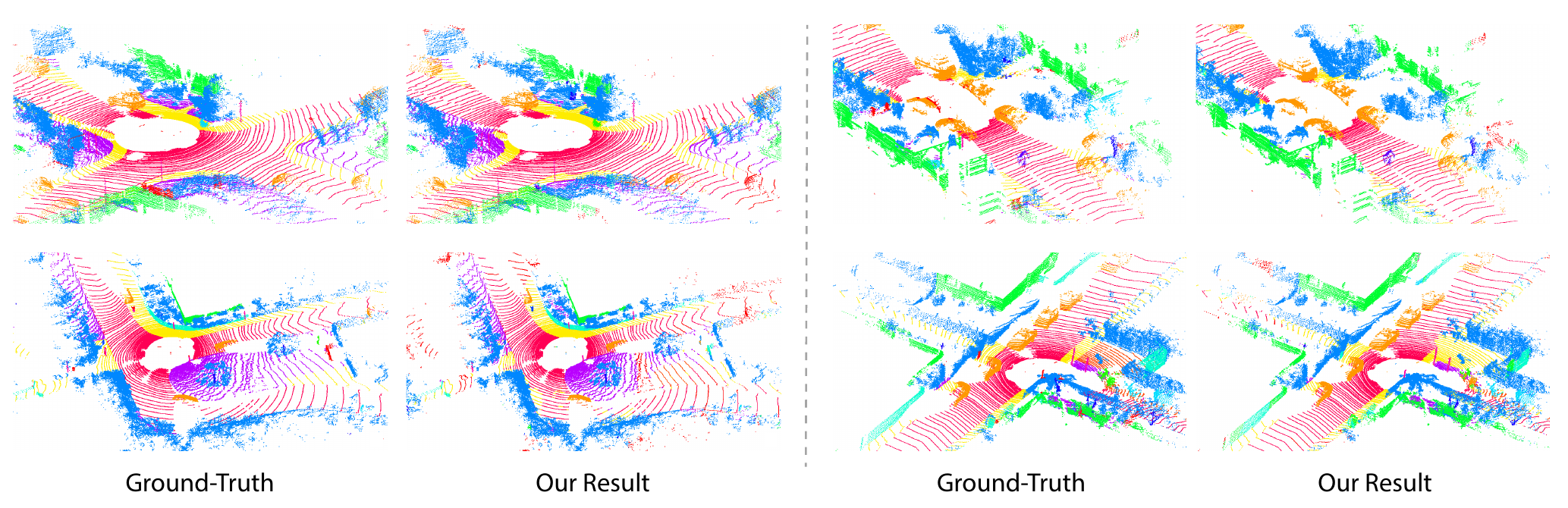}
%\put(3,1){\small (a)}
%\put(20,1){\small (b)}
%\put(37,1){\small (c)}
%\put(55,1){\small (d)}
\end{overpic}
\caption{
A gallery of HPC-DNN segmentation results on the challenging SemanticKITTI dataset.
}
\label{fig:kittigallery}
\end{figure*}

\subsection{Visualization of Results}

We visualize the feature map of our HPC-DNN network after convolution, particularly the extracted features from different kernel shapes. Besides, we visualize the segmentation results of HPC-DNN on S3DIS and SemanticKITTI.

Figure~\ref{fig:FeatureMaps} shows the feature maps of the same layer with different kernels on a S3DIS point cloud scene.
In some feature channels, the features extracted from each kernel has obvious characteristics.
In two different feature channels, the spherical kernel produces similar convolution response on the wall in the same direction(left sub-figure) and in arbitrary directions(right sub-figure), respectively.
It shows that the network has learned directional information by spherical kernel.
Meanwhile, the planar kernel strengthens edge structures.
The vertical linear kernel is robust to the vertical walls with the same feature distribution.
The center point kernel has a local strengthening effect on the point cloud associated with input features.

Figure~\ref{fig:s3disgallery} shows some results of scene segmentation on S3DIS. It can be seen that our method attains accurate segmentation results; it is especially good at distinguishing objects with geometry discrepancy. The segmentation error mainly comes from the semantic objects on the facade with similar planar shape. Figure~\ref{fig:kittigallery} demonstrates the predicted labels by our proposed network on SemanticKITTI. As can be seen, small objects, e.g., vehicle, on road scene are well distinguished by our method.

\section{Discussion and Future Work}

We have presented a method for 3D point cloud convolution using Hausdorff distance between neighborhoods of the query points and a set of kernel points.
The convolution operation is share-aware and allows a powerful feature learning with a small set of geometric priors/kernels.
Based on this geometric convolution operator, we extend conventional regular CNNs to HPC-DNN, using multi-kernels representing different geometric priors.
We further extend our approach and define a weighted Hausdorff distance, which fuses features from adjacent levels and local geometric information in the network.
Evaluations on point cloud scene dataset S3DIS and SemanticKITTI demonstrates that our method is effective, about $2\%$ higher than strong baselines.

Our HPC-DNN has similar limitations as conventional CNNs: the number of kernels, i.e., convolution filters, is limited by the capacity of the deep network.
Furthermore, the kernel points represent prescribed geometric prior. Their shape is not learnable; only their shortest distance weighting is learned.
For extreme sparse point cloud data, the geometry is hard to be extracted, hindering the extraction of effective features by the HPC-DNN.
In the future, we would like to explore different architecture based on HPC, where, in a longer term, explore the possibility to learn the kernel geometries. %during training.
The challenge is to keep the valuable properties of permutation and scale invariance.
%We also call more attentions on how Hausdorff point convolution is used for 3D deep feature learning, e.g., taking each kernel point coordinate value as a network weight. 

{\small
\bibliographystyle{ieee}
\bibliography{points}
}

\end{document}